\def\year{2016}\relax
\documentclass[letterpaper]{article}
\usepackage{aaai16}
\usepackage{times}
\usepackage{helvet}
\usepackage{courier}
\usepackage[linesnumbered,lined,boxed,commentsnumbered]{algorithm2e}
\usepackage{lipsum}
\usepackage{array,amsmath}
\newcolumntype{L}{>{\centering\arraybackslash}m{3cm}}
\usepackage{amsfonts,amssymb,epsfig,epstopdf,url}
\newtheorem{thm}{Theorem}
\newtheorem{defn}[thm]{Definition}

\usepackage{color}

\newcommand{\rf}{\mbox{{\em Rf}}\:}

\begin{document}
 
%
\title{Combining Gradient Boosting Machines\\ with Collective Inference to Predict Continuous Values}
\author{Iman Alodah\\
Computer Science Department\\
Purdue University\\
West Lafayette, Indiana 47906\\
Email: ialodah@purdue.edu\\
\And 
Jennifer Neville\\
Computer Science and Statistics Departments\\
Purdue University\\
West Lafayette, Indiana 47906\\
Email: neville@cs.purdue.edu\\
}
\maketitle
\begin{abstract}
Gradient boosting of regression trees is a competitive procedure for learning predictive models of continuous data that fits the data with an additive non-parametric model. The classic version of gradient boosting assumes that the data is independent and identically distributed. However, relational data with interdependent, linked instances is now common and the dependencies in such data can be exploited to improve predictive performance. Collective inference is one approach to exploit relational correlation patterns and significantly reduce classification error. However, much of the work on collective learning and inference has focused on discrete prediction tasks rather than continuous. 
In this work, we investigate how to combine these two paradigms together to improve regression in relational domains. Specifically, we propose a boosting algorithm for learning a collective inference model that predicts a continuous target variable. In the algorithm, we learn a basic relational model, collectively infer the target values, and then iteratively learn relational models to predict the {\em residuals}. We evaluate our proposed algorithm on a real network dataset and show that it outperforms alternative boosting methods. However, our investigation also revealed that the relational features interact together to produce better predictions.

\end{abstract}
\section{Introduction}
Supervised learning methods use observed data to fit predictive models. The generalization error of these prediction models can be decomposed into two main sources: {\em bias} and {\em variance}. In general, for a single model, there is a tradeoff between bias and variance. However, ensemble methods attempt to resolve this issue by learning a combination of models. {\em Boosting} is an ensemble approach that creates a strong predictor from a combination of multiple weak predictors. The method aims to reduce errors due to bias {\em and} variance to produce better predictions. In this work, we consider the gradient boosting algorithm \cite{Friedman00greedyfunction}, which fits an additive model to independent and identically distributed (i.i.d.) data in a forward stage-wise manner. 

However, many current datasets of interest do not consist of i.i.d. instances, but instead contain structured, relational, network information. For example, in Facebook users are connected to friends, and information about the friends (e.g., demographics, interests) can be helpful to boost prediction results. 
As such, there has been a great deal of recent work focused on the development of {\em collective inference} methods that can exploit relational correlation among class label values of interrelated instances to improve predictions \cite{Chakrabarti:1998:EHC:276305.276332,JNeville:2000,Taskar:2002:DPM:2073876.2073934,Lu:2003a,Macskassy:2007:CND:1248659.1248693,sen:aimag08,McDowell:2009:CCC:1577069.1755879}. Collective inference methods typically learn local relational models and then apply them to {\em jointly} classify unlabeled instances in a network. Since network data often exhibit high correlation among related instances, this results in improved classification performance compared to traditional i.i.d. methods.

In this  paper, we investigate the combination of gradient boosting (GB) and collective inference (CI) to predict continuous-valued target variables in partially-labeled networks. 
While boosting has previously been considered in the context of statistical relational learning \cite{Natarajan10boostingrelational,Khot:2011:LML:2117684.2118261,Natarajan:2012:GBS:2123932.2123937,Hadiji:2015:PDN:2815942.2815988,Khot:2015:GBS:2812521.2812532}, our work is different in two keys aspects. First, we consider the problem of  regression rather than discrete classification. Second, our proposed method uses {\em residuals} at every stage of the gradient boosting algorithm, and uses collective inference to exploit correlations in the the residuals during prediction. 
Specifically, we propose a {\em modified gradient boosting} (MGB) algorithm for learning a collective inference model for relational regression. In the MGB algorithm, we learn a basic relational model, collectively infer the target values, and then iteratively learn relational models to predict the {\em residuals}. We evaluate MGB on a real network dataset and show that it outperforms alternative GB methods. We also conduct some ablation studies that show the importance of the various relational features. The results indicate that relational features involving the neighbors' residuals were not as helpful as features that use the (current) boosted predictions.


The paper is organized as follows. We first review the basic GB and CI methods used to develop our algorithm. Then, we describe the proposed MGB algorithm in detail, including both learning and inference. Next, we describe the experimental setup and show the results. Finally, we discuss related work and conclude. 

\section{Background}
\subsection{Gradient Boosting Machine (GB):}
Gradient boosting machines \cite{Friedman00greedyfunction} are a powerful machine-learning technique that combines a set of weak learners into a single strong learner by fitting an additive model in a forward stage-wise manner. Given a random target variable $Y$ and a set of random attribute variables $X={X_1,X_2,...,X_p}$, the goal is to use the training samples $<{y_i , \mathbf{x}_i}>$ to estimate a function ${F}(\mathbf{x})$ that maps $\mathbf{x}$ to $y$. An estimate of the true underlying function $F^{\star}(\mathbf{x})$ is learned by minimizing the expected value of a specified loss function $L$: 
\begin{equation}
\label{eq:eqstar}
\arg\min_F E_{y,\mathbf{x}}[L(y,F(\mathbf{x}))]
\end{equation}

\noindent The approximation function is of the form:
\begin{equation}
\label{eq:eq1}
F_{M}(\mathbf{X};\rho_{m},\theta)=\sum_{m=0}^{M}\rho_{m}h_m(\mathbf{X};\theta)
\end{equation}

\noindent where $h_m(\mathbf{X};\theta)$ is a function of the input variables $\mathbf{x}$ (parametrized by $\theta$) and $\rho_{m}$ is a multiplier.  $\{h_m(\mathbf{X})\}_{1}^{M}$ are incremental functions (or {\em boosts}) and $h_0$ is the initial guess. 
The goal is to minimize the objective function using a greedy stage-wise approach:
\begin{equation}
\label{eq:eq2}
\left(\rho_{m}, \theta_m \right)= \arg\min_{\rho,\theta}\sum_{i=1}^{N}L\!\left(y_i,\bigr[F_{m-1}(\mathbf{x}_i)+\rho \cdot h(\mathbf{x}_i;\theta)\bigr]\right)
\end{equation}
\noindent where $L$ is the loss function. To solve equation~\ref{eq:eq2} we choose $h(\mathbf{x};\theta)$ that is highly correlated with the negative gradient by solving:
\begin{equation}
\label{eq:eq4}
\theta_m= \arg\min_{\theta}\sum_{i=1}^{N}\bigr[\widehat{y}_i-
h(\mathbf{x}_i;\theta)\bigr]^2
\end{equation}
\noindent where $\widehat{y}_i$ is the residual: 
\begin{equation}
\label{eq:eq5}
\widehat{y}_i= y_i - F(\mathbf{x}_i)  = -\left[\frac{\partial L \bigr(y_i,F(\mathbf{x}_i)\bigr)}{\partial F(\mathbf{x}_i)}\right]_{F(\mathbf{X})=F_{m-1}(\mathbf{X})}
\end{equation}
The multiplier $\rho_m$ is estimated by applying line search:
\begin{equation}
\label{eq:eq6}
\rho_m= \arg\min_{\rho}\sum_{i=1}^{N} L\!\left(y_i,\bigr[F_{m-1}(\mathbf{x}_i)+\rho \cdot h(\mathbf{x}_i;\theta_m)\bigr]\right)
\end{equation}

\noindent At each stage the model is updated as follows:
\begin{equation}
\label{eq:eq3}
F_{m}(\mathbf{X})=F_{m-1}(\mathbf{X})+\rho_{m}h(\mathbf{X};\theta_m)
\end{equation}

\begin{algorithm}[t!]
\SetKwData{Left}{left}\SetKwData{This}{this}\SetKwData{Up}{up}
\SetKwFunction{Union}{Union}\SetKwFunction{FindCompress}{FindCompress}
\SetKwInOut{Input}{Input}\SetKwInOut{Output}{Output}
\Input{data samples $\{<y_i,\mathbf{x}_i>\}_1^N$, iterations $M$}
\Output{Model $F_M$ which has the form of Eq.~\ref{eq:eq1} }

\DontPrintSemicolon
\caption{Gradient Boosting}
\label{algo:gb}
\BlankLine
Compute initial guess $F_0(X)$, e.g., for  least square regression use $F_0=\bar{y}$.\;
\For {m =1 to M}{
			Compute the gradient for each example with Eq.~\ref{eq:eq5}. For regression $\widehat{y}_i=y_i-F_{m-1}(\mathbf{x}_i)$.\; 
			Estimate parameters $\theta_m$ with Eq.~\ref{eq:eq4}. \;
			Compute the step length $\rho_m$ with Eq.~\ref{eq:eq6}.\;
			Update the model $F_{m}$ with Eq.~\ref{eq:eq3}.\;
	}
return $F_M$\;
\end{algorithm}

One choice for the models $h$ are regression trees. In this work, we fit the negative gradient with a regression tree. Algorithm~\ref{algo:gb} shows pseudocode for general gradient boosting. Note that lines 4-5 refer to learning a regression tree model, if regression trees are used as the basic model $h$. $\rho$ is computed for each region in the regression tree as suggested by \cite{Friedman00greedyfunction} and it is actually done by taking the median of the region.

\subsection{Iterative Collective Inference:}
\label{ICAS}
{\em Collective classification} is jointly used to infer the unknown class labels in a partially-labeled attributed graph $G=(V,E)$. Each node $v_i \in V$ has an associated class label and attributes $<\!y_i, \mathbf{x}_i\!>$. In {\em transductive} collective classification, the goal is to learn a model from the subset of nodes in $G$ that are labeled (which we refer to as $Y_k$), and then apply collective inference to make predictions for the remaining (unlabeled) nodes in the graph $Y_u = Y-Y_k$. To exploit the relational information, we use a {\em relational} learner model that is trained using the objects' features and relational features \rf of their neighbors. 

Algorithm~\ref{algo:ica} shows the iterative classification algorithm ICA we use for collective inference. Iterative classification algorithm is one of several algorithms that performs collective classification. 
In this work, we will use the same process to perform iterative collective {\em regression}. In other words, our main focus in this work is to use ICA to predict values $\widetilde{Y}$ in continuous space which uses $\mathcal{N}_i$ the neighbors of node $v_i$ in the prediction process. The relational features that we compute are aggregations of continuous class values (e.g., average and median).

\IncMargin{0.275em}
\begin{algorithm}
\DontPrintSemicolon
\SetKwData{Left}{left}\SetKwData{This}{this}\SetKwData{Up}{up}
\SetKwFunction{Union}{Union}\SetKwFunction{FindCompress}{FindCompress}
\SetKwInOut{Input}{Input}\SetKwInOut{Output}{Output}
\Input{Graph $G=(V\!=\!V_k+V_u,E, \mathbf X, Y_k)$, $F$ \break V contains two types of nodes: $V_k$ have known class labels and $V_{u}$ are unknown; $\mathbf x_i$ are the features of node $v_i$;
; $F$ is the learned model.}
\output{$\widetilde{y}$ Estimations of unknown labels $Y_{u}$}
\caption{Iterative Classification Algorithm ICA}
\label{algo:ica}
\BlankLine
\For {each $v_i$ in $V_u$}{
			Compute relational features ($\rf_{\!i}$) using $v_i$'s observed neighbors (i.e., $y_j \in Y_k \: s.t. \: v_j \in \mathcal{N}_i$) \;
			Construct feature vectors $A_i = [x_i,\rf_{\!i}]$ \;
			$\widetilde{y}_i$ $\leftarrow$ $F(A_i)$
}

\Repeat ( \emph {Iterative inference}) {converges or number of iterations have elapsed}
{
Generate random ordering $O$ over $V_{u}$\;
	\For {each $v_i$ in $O$}{
			Compute relational features ($\rf_{\!i}$) using known labels and current predictions for $\mathcal{N}_i$ (i.e., $y_j \mbox{ for } v_j \in V_k$ and $\widetilde{y}_j \mbox{ for } v_j \in V_u$) \;
			Construct feature vectors $A_i = [x_i,\rf_{\!i}]$ \;
			$\widetilde{y}_i$ $\leftarrow$ $F(A_i)$
	}
}
()
\end{algorithm}
\DecMargin{0.275em}

\section{Collective Inference Gradient Boosting \break For Continuous Variables}
\subsection{Problem Statement:}
Many relational and social network datasets exhibit correlation among the class labels values of linked nodes. While there is a large body of work focused on learning collective classification models for predicting discrete target variables, there is relatively less work focused on continuous class labels. In this paper we focus on developing a boosting algorithm for predicting a continuous target variable $Y$.
\begin{defn} Transductive collective classification in relation data with continuous class labels: \end{defn}
\vspace{-1mm}
\noindent Given a partially labeled graph $G=(V,E,\mathbf{X},Y)$ where $V$ is a set of nodes, $E$ is a set of edges, $\mathbf{X}$ is set of features, and $y_i \in Y$ is a continuous target value for each node $v_i \in V$. The nodes $V$ are split into a {\em known} set $V_k$ with labels $Y_k$ and an {\em unknown} set $V_u$ s.t., $V=V_k+V_u$. The goal is to learn a collective classification model from the induced subgraph over $V_k$ and apply it to $G$ to predict class label values for $Y_u$.

\subsection{Model:}
In this work, we develop a gradient boosting algorithm that uses collective inference (CI) in the learning phase. Since we focus on regression, we modify the gradient boosting algorithm that minimizes squared loss as described in Friedman \shortcite{Friedman00greedyfunction} and combine it with with CI. Our Modified Gradient Boosting (MGB) algorithm is described in Alg.~\ref{algo:mgb}. The algorithm fits a stronger non-parametric function as in Eq.~\ref{eq:eq1}---the learning is guided by CI over the continuous target variable $Y$, which uses the full model $F_m$ and the residuals $\widehat{y}_i$ that are predicted using the weak learned models $\{h\}_1^m$.  In the application phase, we use CI to collectively infer the values of $Y_u$ using the induced relational gradient-boosted model. Algorithm \ref{algo:ica2} shows how we apply the ICA algorithm in our setting. Note we use $\widehat{y}_i$ refer to the residual of the node $v_i$ and $\widetilde{y_i}$ refer to the estimation of the class label of the node $v_i$ (i.e., $\widehat{y}_i = y_i - \widetilde{y_i}$).

The base model of our MGB algorithm is a regression tree. Our model is an additive model that learns a tree at each stage, and then the final result is the sum of the predictions from each tree. We made one additional modification to the original squared loss gradient boosting regression in Friedman \shortcite{Friedman00greedyfunction}.  Since our data exhibits a skewed class distribution, we use the median instead of the average to compute predictions in the terminal nodes of each tree. Using the median prevents the predictions from being skewed towards the extreme values in the distribution. 

\IncMargin{0.275em}
\begin{algorithm}[h!]
\SetKwData{Left}{left}\SetKwData{This}{this}\SetKwData{Up}{up}
\SetKwFunction{Union}{Union}\SetKwFunction{FindCompress}{FindCompress}
\SetKwInOut{Input}{Input}\SetKwInOut{Output}{Output}
\Input{Graph $G_{tr}=(V_k,E_k, \mathbf X_k, Y_k)$, the induced subgraph over nodes with known labels; \\ 
$l$: max number of nodes in regression tree; \\
$M$: number of boosting iterations;\\
$t$: number of $\widetilde{Y}$ estimations.
\\}
\Output{Model $F_M$ which has the form of Eq.~\ref{eq:eq1}. }
\DontPrintSemicolon
\caption{Modified Gradient Boosting (MGB)}
\label{algo:mgb}
Compute relational features $\rf$ using $Y_k$ \;
Construct feature vectors $\mathbf A_0= [\mathbf X_k,\rf]$ $\forall v \in V_k$ \;
Fit regression tree $F_0(\mathbf A_0)= h_0(\mathbf A_0,l)$ to predict $Y_k$ \;
\For { t iterations}{
Randomly split $G_{tr}$ into 80\% to use for $Y_{tk}$  and 20\% for $Y_{tu}$. \;
$\widetilde{Y}_{tu}$ = ICA2($V_{tk},V_{tu},E_k, \mathbf X_k, Y_{tk},F_0$) \;
}
$\widetilde{y_i}$= average of $\widetilde{y}_{ti}$ for $t \in {1..t}$\;

\For {m =1 to M}{
			Compute residuals $\widehat{y}_i=y_i-\widetilde{y_i}$ for all $v \in V_k$ \;
			Compute relational features $\rf$ using $Y_k, \widehat Y_k, \widetilde Y_k$ \;
			Construct feature vectors $\mathbf A_m= [\mathbf X_k,\rf]$ \;
			Fit regression tree $h_m(\mathbf A_m,l)$ to predict $\widehat{Y}$\;
			Update the final model: $F_m(\mathbf A_{[0..m]}) = F_{m-1}(\mathbf A_{[0..m-1]})+ h_m(\mathbf A_m)$\;
			
			\For { t iterations}{
			Randomly split $G_{tr}$ into 80/20\% for $Y_{tk}$ / $Y_{tu}$ \;
			$\widetilde{Y}_{tu}$ = ICA2($V_{tk},V_{tu},E_k, \mathbf X_k, Y_{tk},F_m$) \;
			}
			$\widetilde{y}_i$= average of $\widetilde{y}_{ti}$ for $t \in {1..t}$\;
	}
return $F_M$\;
\end{algorithm}
\DecMargin{0.275em}
\setlength{\textfloatsep}{0pt}

\IncMargin{0.275em}
\begin{algorithm}[h!]
\DontPrintSemicolon
\SetKwData{Left}{left}\SetKwData{This}{this}\SetKwData{Up}{up}
\SetKwFunction{Union}{Union}\SetKwFunction{FindCompress}{FindCompress}
\SetKwInOut{Input}{Input}\SetKwInOut{Output}{Output}
\Input{Graph $G\!=\!(V\!=\!V_k\!+\!V_u,E, \mathbf X, Y_k)$, $F_M$, $itr$ \break 
$F_M$ is the learned model; \\
$itr$: number CI iterations.}
\Output{$\widetilde{y}\;$ estimations of unknown labels $Y_{u}$}
\caption{ICA2 for MGB }
\label{algo:ica2}
\BlankLine
 {\scriptsize \texttt{\# compute residuals of known set}} \break
\For {$m \in [1,M]$ }{
\For {each $v_i$ in $V_k$}{
			Compute relational features ($\rf_{\!im}$) using $v_i$'s observed neighbors (i.e., $y_j \in Y_k \: s.t. \: v_j \in \mathcal{N}_i$) \;
			Construct feature vector $A_{im} = [x_i,\rf_{\!im}]$ \;
			$\widetilde{y}_{im}$ $\leftarrow$ $F_{m-1}(A_{im})$ \;
			Compute true residual $\widehat{y}_{im}=y_i- \widetilde{y}_{im}$
}
}

 {\scriptsize \texttt{\# initialize predicted class/residuals for unknown}} \break
$\mathbf{\mathcal{Y}}_i \leftarrow [ \; ] \;\;\forall  v_i \in V_u$ \;
\For {$m \in [0,M]$}{
\For {each $v_i$ in $V_u$}{
			Compute relational features ($\rf_{\!im}$) using $v_i$'s observed neighbors' class and residuals (i.e., $[y_j,\{\widehat{y}_{im}\}_1^M]  \: s.t. \: y_j \in Y_k \wedge \: v_j \in \mathcal{N}_i$) \;
			Construct feature vector $A_{im} = [x_i,\rf_{\!im}]$ \;
			\textbf{if} $m=0$ \textbf{then} $\widetilde{y}_i \leftarrow h_{m}(A_{im})$; $\mathcal{Y}_{i}[m] = \widetilde{y}_i$\;
			\textbf{else} $\widehat{y}_{im} \leftarrow  h_{m}(A_{im})$; $\mathcal{Y}_{i}[m] = \widehat{y}_{im}$ 
}
}
 {\scriptsize \texttt{\# collective inference over unknown set}} \break
\For {$t \in [0,itr]$}{
Generate random ordering $O$ over $V_{u}$\;
	\For {each $v_i$ in $O$}{
		\For {$m \in [0,M]$}{
			Compute relational features ($\rf_{\!im}$) using known labels/residuals and current predictions 
			(i.e., $[y_j,\{\widehat{y}_{im}\}_1^M]  \mbox{ for } v_j \!\in\! V_k$ and $\mathcal{Y}_j \mbox{ for } v_j \in V_u$) \;
			Construct feature vector $A_{im} = [x_i,\rf_{\!im}]$ \;
			\textbf{if} $m=0$ \textbf{then} $\widetilde{y}_i \leftarrow h_{m}(A_{im})$; $\mathcal{Y}_{i}[m] = \widetilde{y}_i$\;
			\textbf{else} $\widehat{y}_{im} \leftarrow  h_{m}(A_{im})$; $\mathcal{Y}_{i}[m] = \widehat{y}_{im}$ 
	}
}
}
\Return $\sum_0^m \mathcal{Y}_{im}  \;\;\forall  v_i \in V_u$ 

\end{algorithm}
\DecMargin{0.275em}

\setlength{\textfloatsep}{0pt}
Algorithm~\ref{algo:mgb} starts by fitting a regression tree with a specified depth $l$. Then the input graph is divided into two separate sets for CI: 20\% are set as unknown in $Y_{tu}$ and 80\% are set as known in $Y_{tk}$. The division of labeled/unlabeled can be tuned experimentally. These two sets are then used in the CI process when ICA2 is applied. We repeat the CI multiple times to ensure that every node in the input graph is set as unknown at least once. This can be achieved by choosing a random order of $G_{tr}$ at every iteration then dividing it into the two groups as discussed above. The multiple predictions for each node are averaged afterwards (line 8).

The algorithm then enters the loop that fits a set of models that constitute the {\em residual} models. In line 10, the algorithm computes the residuals (i.e., the negative gradient). In line 11, the relational features are updated. Note that each model in $\{F_m\}_0^m$ has a potentially different set of relational features. The features are specific to each model $F_m$  because they depend on the target value of that model (i.e., class label or residual). 
The next section describes the set of relational features in more detail.

Once the relational features are recalculated, the algorithm learns a new tree model over the residuals then it applies ICA2 with the new complete model $F_{m-1} +h$, where h is the regression tree learned on the current iteration. This approach aims to exploit the correlation in the residuals for more accurate predictions. As before, we need to repeat the inference process multiple times to ensure that each node gets at least one prediction from CI. The predictions are then averaged to be used in the next iteration of residual calculation. The number of specified iterations/models can also be tuned experimentally using a validation set to prevent overfitting. 

ICA in algorithm~\ref{algo:ica2} predicts not only the class label but also the residuals predicted by each sub model. The first loop from line 1 to 8 computes the true residuals for the known nodes. This information is need in the next loop from line 10 to 17 where we compute initial prediction for all the unknown nodes. Finally,  in line 18 to 28 we apply collective inference to predict the class label and the residuals. Out final class label prediction is the sum of all the class label prediction and the residuals predicted in line 18 to 28.

\subsection{Features:}
\label{feats}
Along with non-relational features, we use four different relational features. 
The first relational feature $\rf_{\!1}$ records the median of the neighbors' target values. The initial value of this feature is the median of the target values of the known neighbors and during the collective inference it is calculated using the known values and the estimated values over the neighbors:
\begin{equation}
\label{eq:eq8}
\rf_{\!1j}= median\{\mathcal{N}_j(Y)\} 
\end{equation}
where $Y$ are  known or estimated values for the neighbors ($\mathcal{N}_j$) of node $v_j$.

 The features $\{\rf_{\!2m}\}_0^M$ record the median of the neighbors' target values at each boosting stage $m$. The initial model $F_0$ predicts the class, so the feature will be the median of the target values. The models $m={1... M}$ predict the residuals, so the associated feature values will be the median of the residuals computed at that stage: 
\begin{equation}
\rf_{\!2mj}= median\{\mathcal{N}_j(\widehat{Y}_{m})\} 
\end{equation}
Here $\widehat{Y}_{m}$ refers to the known or estimated residuals at stage/model $m$. Note that at stage $0$ we can view the full target value as the residual if we consider the initial estimation as 0. 

The features $\{\rf_{\!3m}\}_0^M$ record the median of the residuals of the boosting stage $m-1$, but it is used to train the model at stage $m$.
\begin{equation}
\rf_{\!3mj}= median\{\mathcal{N}_j(\widehat{Y}_{m-1})\} 
\end{equation}
Here $\widehat{Y}_{m-1}$ are the known or estimated residuals at the stage/model $m-1$.

Finally, the feature $\rf_{\!4j}$ records the instance $v_j$'s target value estimated by the complete model so far (i.e., $F_{m-1}$). This feature is used in stage $m$ to learn a new model: 
\begin{equation}
\rf_{\!4j}= F_{m-1}(A_j)
\end{equation}

\section{Experimental Evaluation}
\subsection{Hypothesis:}
Our hypothesis is that, when using gradient boosting to learn a relational regression model, exploiting relational correlation---of both predicted target values and residuals---can outperform learning using fixed relational features. In other words, we want to answer the following question:\\
Q1: Does using collective inference in the learning phase of the gradient boosting improves its prediction of continuous values compared to both gradient boosting without any relational features (GB) and gradient boosting with relational features but no collective inference (RGB)?

\subsection{Datasets:}
We evaluated our proposed algorithm on the IMDB data. We considered a network of movies that are linked if they share a producer, and the target label is the movie's total revenue. The largest connected component of this dataset has about 10770 nodes. Each movie has 29 genre features and we added the four relational features discussed above. Some statistics of the dataset are shown in Table~\ref{table:2}. 

\subsection{Baseline Models:}
We compare our proposed MGB model to two different baselines. The first natural choice is the gradient boosting machine (GB) as proposed in Friedman~\shortcite{Friedman00greedyfunction}, which does not exploit the relational nature of network data. We use the 29 non-relational features to learn the model. The second baseline is a simple relational version of the gradient boosting machine (RGB) which is described next.
\subsubsection{Relational Gradient Boosting:}
\label{RGB}
The RGB model is trained as regression tree with gradient boosting---using additional relational features. Along with the 29 non-relational features that we use to learn GB, we add one more relational feature that records the median of the known neighbors' target labels $Y_k$. This is a simple relational model in which the attributes of an object depend on the class label of that object as well as the target labels of objects one link away \cite{Jensen04whycollective}. The relational feature is computed as described in Eq.~\ref{eq:eq8}.

\begin{table}
\caption{IMDB Dataset Characteristics}
\label{table:2}
\begin{tabular}{ | p{5cm} | p{2.5cm} | }
  \cline{1-2}
  \multicolumn{2}{|c|}{IMDB }\\
  \hline
   Characteristic&  Value  \\
  \hline 
  Number of nodes &  10770\\
 \hline 
  Number of edges &  212460 \\
 \hline 
  Labels correlation coefficient &  0.22699 \\
 \hline 
  Average degree &  19.727 \\
 \hline 
  Average clustering coefficient &  0.53947 \\
 \hline 
   Number of non-relational features &  29 \\
 \hline
   Number of relational features &  4 \\
 \hline
  Target label & Continuous value from 30 to $1054\times10^6$  \\
 \hline
\end{tabular}
\vspace{3mm}
\end{table}

\subsection{Experimental setup:}

\subsubsection{Training/Test Splits Generation}
We divide the network into 5 folds, where each is 20\% of the data. Then we set the training set to be $\{20\%,40\%,60\%,80\%\}$ of the data and the testing set to (100\% - training set). 

\subsubsection{Test Procedure}
We use 5-fold cross-validation for evaluation. The learning phase of the algorithms is implemented as described in Algorithm~\ref{algo:mgb}. For prediction, we apply the learned model using the ICA collective inference algorithm for RGB and the ICA2 collective inference algorithm for MGB. The accuracy of the predictions are evaluated with the root mean square error (RMSE).

The algorithms have different parameters that need to be specified in advance. For the tree-based gradient algorithm, we have two main parameters: (1) The number of boosting iterations M---which we set to 10, and (2) the max size of the regression tree $l$---which we set to 5. Both parameters can be tuned better by using a validation set and avoid over fitting the training set, but we leave this to future work. 

We have another parameter that is related to collective inference. From the literature of iterative collective classification, 50 iteration ($itr$) are sufficient.  Since we have continuous values instead of discrete, we assessed the convergence of the estimations over the iterations and found that the estimations converge at a rate similar to discrete classification. Thus, we set $itr=50$.

Our MGB algorithm has two additional parameters. The number of times $t$ that inference is repeated to get an estimate for all the nodes. We set $t=3$. Moreover we randomly selected 20\% (80\%) of the training set for the unknown (known) subset for each trial $t$.

\subsection{Results and Discussion:}
\begin{figure}[]
\centering
                     \includegraphics[width=0.85\columnwidth]{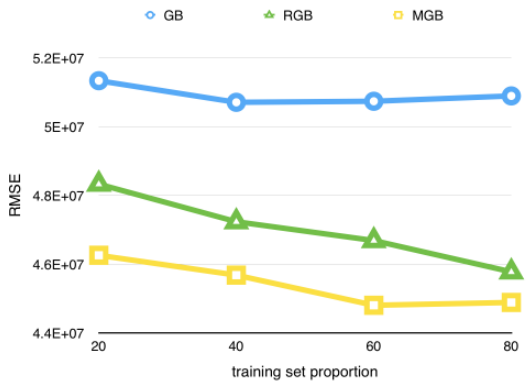}
\vspace{-2mm}                 
                    \caption{Performance of GB, RGB, and MGB on IMDB.}
                    \label{perfor}
\vspace{2mm}                 
\end{figure}

Our experiments show that incorporating collective inference in the learning phase improves performance---since the continuous target  labels are correlated and the residuals (error) are also correlated. Figure~\ref{perfor} shows how our approach (MGB) outperforms the the baseline algorithms. Our algorithm achieves lower RMSE values when tested using different training set sizes (i.e., data proportion). We test the algorithms first by training using a random 20\% of the network and apply the induced model on the remaining 80\%. We then increase the proportions of the network allocated to training through to 80\%. The performance of the MGB algorithm is consistently better than the other methods. The results also show that a larger portion of the network is labeled, the performance increases for all models (lower RMSE value) and the added gain provided by collective inference decreases.

\begin{figure}[]
 \centering
\vspace{-2mm}                 
                    \includegraphics[width=0.85\columnwidth ]{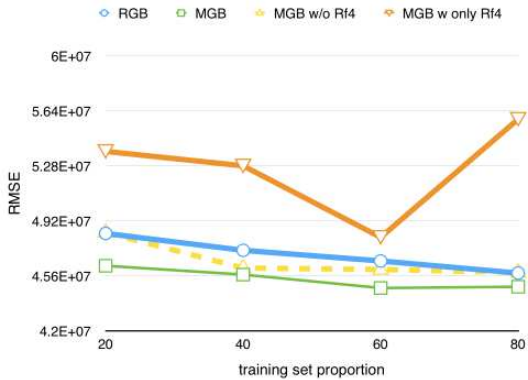}
                    \caption{Performance of MGB learned using only $\rf_{\!4}$ and MGB learned with all the features but  $\rf_{\!4}$ (on IMDB).}
                    \label{comparefeats}
\vspace{2mm}                 
\end{figure}

\begin{figure}[]
 \centering
\vspace{-2mm}                 
                     \includegraphics[width=0.85\columnwidth]{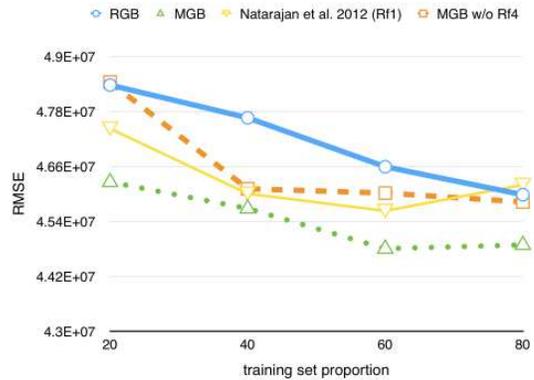}
                    \caption{Performance of MGB compared to implementation of Natarajan et al. \shortcite{Natarajan:2012:GBS:2123932.2123937} for regression (on IMDB).}
\vspace{4mm}                 
                    \label{prevwork}
\end{figure}

One of the main factors that affect the performance is autocorrelation between the labels, the autocorrelation between the residuals and finally the correlation between the residuals and the true or estimated target values.  In IMDB dataset, the labels are autocorrelated and the residuals are also autocorrelated. Feature $\rf_1$ and feature $\rf_4$ are the most effective features of the four suggested features. Feature Rf1 utilizes the autocorrelation in the target label and using it alone gives good results. Using feature $\rf_1$ alone is comparable to how we would interpret the work of  \cite{Natarajan:2012:GBS:2123932.2123937} for regression, note that both their task and the type of the target variable are different than ours. As for feature $\rf_4$, it exploits the correlation between the residuals and the estimated target value. Using this feature alone gives good results as well, however, combining them gives the best performance. Figure ~\ref{comparefeats} shows the performance of the MGB with only the relational feature $\rf_4$ and without it. The figure gives good indication that both $\rf_4$ and $\rf_1$ are needed for better performance. As for the relational feature $\rf_1$, Figure ~\ref{prevwork} shows performance comparison when we implement MGB with only $\rf_1$, all feature(MGB) and MGB without $\rf_4$. It is also important to note that both features $\rf_2$ and $\rf_3$ are computed using the residuals. The performance of MGB does not get better when using only these features and might get even worse when rely on these features alone for prediction.

The level of relational dependence impacts performance, including (1) correlation between the labels of linked nodes, (2) correlation between the residuals of linked nodes, and (3) correlation between the residuals and the true or estimated target values.  In the IMDB data, the labels and the residuals are correlated across links. Features $\rf_{\!1}$ and $\rf_{\!4}$ are the most effective features of the four suggested features. Feature $\rf_{\!1}$ utilizes the relation correlation in the target label and using it alone gives good results. Using feature $\rf_1$ alone is comparable to how the Natarajan et al.~\cite{Natarajan:2012:GBS:2123932.2123937} method would be implemented for regression. Note that in their paper, both their tasks and the type of the target variable are different from ours. 

As for feature $\rf_{\!4}$, it exploits the correlation between the residuals and the estimated target value. Using this feature alone produces good results as well, however, combining them results in the best performance. Figure~\ref{comparefeats} shows MGB performance with only the relational feature $\rf_{\!4}$ and without it. The results indicates that both $\rf_{\!4}$ and $\rf_{\!1}$ are needed to achieve best performance. 

As for the relational feature $\rf_{\!1}$, Figure~\ref{prevwork} compares performance when we implement MGB with only the $\rf_{\!1}$ feature, all features (MGB), and MGB without $\rf_{\!4}$. It is important to note that both features $\rf_{\!2}$ and $\rf_{\!3}$ are computed using the residuals. The performance of MGB does not improve when using only these features and might even degrade when rely on these features alone for prediction. Although the error is correlated with the residuals in our dataset, it is difficult to estimate the residuals using collective inference.

The accuracy of the algorithm is sensitive to the manner in which the features are initialized for learning in each model. We set the initial relational features for the models to be equal to the median of the known neighbors' target values for $\rf_{\!1}$ and $\rf_{\!4}$ and zero for $\rf_{\!2}$ and $\rf_{\!3}$. If a given node does not have any known neighbors, all the features are set to zero. 

\section{Related Work}
Boosting for statistical relational learning (SRL) has been studied in previous work such as \cite{Natarajan:2012:GBS:2123932.2123937} and \cite{Natarajan10boostingrelational}. Natarajan et al.~\shortcite{Natarajan:2012:GBS:2123932.2123937} uses gradient boosting to learn relational and non-parametric functions that approximate a joint probability distribution over multiple random variables in {\em relational dependency networks} (RDNs). The algorithm learns the structure and parameters of the RDN models simultaneously.  Since RDNs can be used for collective classification, Natarajan et al. include other query predicates in the training data while learning the model for the current query and then apply the learned mode collectively. The authors compare the boosted RDNs (RDN-B) against Markov logic networks \cite{richardson2006markov} and basic RDNs on two kinds of tasks: entity resolution and information extraction.  The performance of RDN-B was significantly better compared to RDN and MLN for most datasets.  However, the authors don't report a comparison between using RDN-B with and without collective inference. 

Hadiji et al.~\shortcite{Hadiji:2015:PDN:2815942.2815988} introduce non-parametric boosted Poisson Dependency Networks (PDNs) using gradient boosting. Among other objectives in their work, the authors performed collective prediction, however, they again did not compare performance with and without collective inference. 

In Khot et al.~\shortcite{Khot:2015:GBS:2812521.2812532,Khot:2011:LML:2117684.2118261}, the authors used gradient-based boosting for MLNs. The authors derived functional gradients for the pseudo-likelihood and the boosted MLNs outperformed the non-boosted version. However, the boosted models were not used for collective inference. 

We note that all the related work above focused on classification tasks with discrete labels rather than regression.
\section{Conclusion}
In this work we investigated the use of boosting to learn a collective regression model. Our results show that gradient boosting, combined with collective inference, results in improves performance compared to gradient boosting in a relational model that does not use collective inference. This indicates that using the {\em residuals} in the relational model, and including CI in the boosting process, can produce better predictions. 

The work reported in this paper is different from other research on using boosting in SRL methods in several ways. First, our MGB method performs regression instead of classification. Second, we compared the performance of gradient boosting using simultaneous statistical estimates about the same variables for a set of related data instances (i.e., in collective inference) with the classic version of gradient boosting that considers a fixed set of features. We also compared to a simple relational model that does not use CI in the learning phase. Finally, we show that using collective inference to estimate the residuals is difficult and that the most effective features are the ones that does not use neighbors' residuals, but instead use their current boosted predictions.


\section{Acknowledgments}
This research is supported by NSF contract number IIS-1149789 and NSF Science \& Technology Center grant CCF-0939370. 
The U.S. Government is authorized to reproduce and distribute reprints for governmental purposes notwithstanding any copyright notation hereon. 

\bibliographystyle{aaai}
\bibliography{workshop}
\end{document}